\documentclass[runningheads]{llncs}
\usepackage[T1]{fontenc}
\usepackage{graphicx}
\usepackage{booktabs}
\usepackage[misc]{ifsym}
\newcommand{\corr}{(\Letter)}
\usepackage{mwe}
\usepackage{amsmath}
\usepackage{algorithm} 
\usepackage{algpseudocode}
\usepackage{graphicx}
\usepackage{multirow}
\usepackage{hyperref}

\newcommand{\DotPunct}{\mathpunct{\raisebox{0.5ex}{.}}}
\newcommand{\algname}[1]{\textsc{#1}}
\newcommand{\dataset}[1]{\textsc{#1}}

\begin{document}

\title{\algname{FeDABoost}: Fairness Aware Federated Learning with Adaptive Boosting\thanks{This research was funded partly by the Knowledge Foundation, Sweden, through the Human-Centered Intelligent Realities (HINTS) Profile Project (contract 20220068).}}

\titlerunning{\algname{FeDABoost}}

\author{Tharuka Kasthuri Arachchige\corr \and
Veselka Boeva\and
Shahrooz Abghari}

\authorrunning{T. Kasthuri Arachchige, V. Boeva and S. Abghari}

\institute{Department of Computer Science, Blekinge Institute of Technology, Sweden\\ \email{\{tak, vbx, sab\}@bth.se}
}

\maketitle              
 
\begin{abstract}

This work focuses on improving the performance and fairness of Federated Learning (FL) in non-IID settings by enhancing model aggregation and boosting the training of underperforming clients. We propose \algname{FeDABoost}, a novel FL framework that integrates a dynamic boosting mechanism and an adaptive gradient aggregation strategy. Inspired by the weighting mechanism of the Multiclass AdaBoost (\algname{SAMME}) algorithm, our aggregation method assigns higher weights to clients with lower local error rates, thereby promoting more reliable contributions to the global model. In parallel, \algname{FeDABoost} dynamically boosts underperforming clients by adjusting the focal loss focusing parameter, emphasizing hard-to-classify examples during local training. These mechanisms work together to enhance the global model’s fairness by reducing disparities in client performance and encouraging fair participation. We have evaluated \algname{FeDABoost} on three benchmark datasets: MNIST, FEMNIST, and CIFAR10, and compared its performance with those of \algname{FedAvg} and \algname{Ditto}. The results show that \algname{FeDABoost} achieves improved fairness and competitive performance. The \algname{FeDABoost} code and the experimental results are available at \href{https://github.com/TharukaCkasthuri/FeDABoost}{GitHub}

\end{abstract}

\section{Introduction}

Federated learning (FL) is a solution to the problem of centralized learning, which requires a large amount of data and causes privacy, security, and computational challenges. The fundamental idea of FL is decentralized learning, which does not require sending user data to a central server. As an emerging technique, FL effectively addresses the challenge of preserving data privacy by keeping data localized on devices and sharing only model updates rather than raw data to train a global model collaboratively. Techniques such as encryption and secure aggregation can further enhance the privacy of these updates during data transmission~\cite{kaur2023federated}.

FL faces several critical challenges, particularly when working with non-IID (non-Independent and Identically Distributed) data. While traditional aggregation techniques, such as \algname{FedAvg}~\cite{mcmahan2017communication}, are effective in IID settings, they often perform poorly in non-IID environments~\cite{li2019convergence}. The non-IID nature of client data, which can include class imbalances or unique challenging examples for each client, results in variations in the quality of local model updates. This variability affects the global model's ability to generalize effectively~\cite{li2019convergence}. Furthermore, ensuring fairness in client participation during the FL process is a critical issue~\cite{Shi2024}. Current methods often favor clients with more powerful resources, higher data quality, or faster response times. These methods unintentionally marginalize clients with fewer resources and result in biased global models. Furthermore, fairness issues can arise in sharing incentives, as this may overlook unequal contributions from clients, potentially discouraging their future participation.

To improve fairness across clients while maintaining reasonable overall performance in non-IID FL settings, we propose \algname{FeDABoost}, a dynamic boosting-based FL algorithm. The key goal of \algname{FeDABoost} is to reduce disparities in model performance across clients while keeping the average predictive accuracy high. In \algname{FedABoost}, at each round, after receiving the global model, each client evaluates it on their local data, and this feedback is used to dynamically boost the influence of underperforming clients in subsequent rounds. This boosting is achieved by adaptively tuning the focal loss focusing parameter to emphasize hard-to-classify examples. In parallel, \algname{FeDABoost} employs a novel weighting mechanism that assigns higher aggregation weights to clients whose local performance is strong, allowing the global model to better leverage reliable updates. Together, these mechanisms promote fairness by mitigating the effects of data heterogeneity and client imbalance, without resorting to personalized models. Experimental results on MNIST, FEMNIST, and CIFAR10 datasets show that \algname{FeDABoost} achieves improved fairness and competitive performance compared to those of \algname{FedAvg} and \algname{Ditto}.

\section{Problem Formulation}
\label{sec:problem}

Let \( k \) denote the total number of clients participating in the FL setup. Each client \( i \), for $i=1, 2,\ldots, k$, owns a local dataset \( \mathcal{D}_i \), which follows a distinct data distribution.

Clients collaboratively train a global model \(\mathcal{M}\) without sharing raw data. The \(\mathcal{M}\) is evaluated on each client's holdout test set (sampled from \( \mathcal{D}_i \)) using loss $\ell_i(\mathcal{M})$ and performance measure \(\varphi_i(\mathcal{M})\), where $\varphi_i$, e.g., can be F1 score or accuracy. We define the average loss~\eqref{eq:average_loss}:

\begin{equation}
\bar{\ell}(\mathcal{M}) = \frac{1}{k} \sum_{i=1}^{k} \ell_i(\mathcal{M}).
\label{eq:average_loss}
\end{equation}
 and the variance of performance measure~\eqref{eq:fairness_variance}:
\begin{equation}
\text{Var}(\varphi(\mathcal{M})) = \frac{1}{k} \sum_{i=1}^k \left( \varphi_i(\mathcal{M}) - \bar{\varphi}(\mathcal{M}) \right)^2, \quad \text{where } \bar{\varphi}(\mathcal{M}) = \frac{1}{k} \sum_{j=1}^k \varphi_j(\mathcal{M}).
\label{eq:fairness_variance}
\end{equation}

\noindent
A lower \(\text{Var}(\varphi(\mathcal{M}))\) indicates that the model performs more uniformly across all clients. Therefore, given two models $\mathcal{M}$ and $\mathcal{M}'$, if \(\text{Var}(\varphi(\mathcal{M})) < \text{Var}(\varphi(\mathcal{M}'))\), then \(\mathcal{M}\) is considered to be fairer than \(\mathcal{M}'\)~\cite{li2021Ditto}.

Our aim is to develop an improved model \(\mathcal{M}\) that significantly reduces both \(\bar{\ell}(\mathcal{M})\) and \(\text{Var}(\varphi(\mathcal{M}))\) compared to a given baseline \(\mathcal{M'}\), i.e.,

\begin{equation}
\bar{\ell}(\mathcal{M}) < \bar{\ell}(\mathcal{M}') \quad \text{and} \quad \mathrm{Var}(\varphi(\mathcal{M})) < \mathrm{Var}(\varphi(\mathcal{M}')).
\label{eq:objective}
\end{equation}

\section{Background}
\subsection{Multi-class Adaptive Boosting}
\label{sec:adaboost}

Adaptive Boosting~\cite{freund1997decision}, known as \algname{AdaBoost}, is an ensemble learning method originally developed for binary classification. It builds a strong classifier by sequentially training weak learners, increasing focus on misclassified instances in each iteration. The final prediction is a weighted vote of these weak learners.

\algname{SAMME} (Stagewise Additive Modeling using a Multi-class Exponential loss function)~\cite{hastie2009multi}, extends \algname{AdaBoost} to multi-class classification problems. The \algname{SAMME} algorithm retains the core principles of \algname{AdaBoost} but modifies the weight update factor ($\alpha$), by incorporating the number of classes to ensure proper adaptation to the multi-class setting, as shown in~\eqref{Eq:weight-update-factor}.

\begin{equation}
\label{Eq:weight-update-factor}
\alpha_{l} = \ln\left(\frac{1 - \mathcal{E}_l}{\mathcal{E}_l}\right) + \ln\left(C - 1\right),
\end{equation} 

\noindent where, \(C\) represents the number of classes, and \(\mathcal{E}_l\) denotes the weighted classification error of the \(l\)th classifier, calculated as \(\mathcal{E}_l = \sum_{i=1}^{N} w_i I(T_l(x_i) \neq y_i)\). The indicator function \(I(.)\) equals 1 if the sample $x_i$ is misclassified and 0 otherwise. 

To ensure $\alpha_l > 0$, the weak classifier must perform better than random guessing, i.e., $(1 - \mathcal{E}_l) > {1}/{C}$. This requirement is critical in boosting, as weak classifiers that perform worse than random chance would otherwise negatively impact the ensemble model. Additionally, \algname{SAMME} combines weak classifiers slightly different from \algname{AdaBoost} by incorporating \(\log(C - 1)\), expressed as \(\log(C - 1) \sum_{i=1}^{N} I(T(l)(x_i) = c)\). When \(C = 2\), \algname{SAMME} behaves similarly to \algname{AdaBoost}.

\subsection{Focal Loss for Challenging Cases}
\label{sec:focal_loss}

Focal loss~\cite{lin2017focal} was originally proposed to address the class imbalance in object detection by modifying the standard cross-entropy loss to emphasize hard-to-classify examples and reduce the influence of easy ones. The formula of focal loss is given by~\eqref{Eq:focal_loss}:

\begin{equation}
\label{Eq:focal_loss}
    \mathcal{L}_{\text{Focal}}(p_t) = -\beta(1 - p_{t})^{\gamma} \log(p_{t}),
\end{equation}

\noindent where $p_{t}$ is the predicted probability assigned to the ground-truth class, $\beta$ is a balancing factor, and $\gamma \geq 0$ is the focusing parameter. The modulating factor \((1-p_t)^\gamma\) dynamically scales the loss based on the prediction confidence; when \(p_t\) is high (i.e., the prediction is correct and confident), the factor approaches zero, reducing the loss contribution from easy examples. Conversely, for hard or misclassified examples where \(p_t\) is low, the modulating factor remains near one, preserving a high loss and encouraging the model to focus on these samples. Increasing \(\gamma\) amplifies this effect, further prioritizing difficult examples during training.

\section{Methodology}

\subsection{Aggregation Mechanism in \algname{FeDABoost}}
\label{sec:aggregation}

Inspired by \algname{SAMME}, we adapt the weight update factor ($\alpha$), originally defined in~\eqref{Eq:weight-update-factor}, for the FL setup. In \algname{SAMME}, weak learners (weak classifiers) are trained sequentially, with each iteration adjusting the sample weights based on previous errors. However, in FL, clients train independently and in parallel~\cite{mcmahan2017communication}, making sequential re-weighting infeasible. 

Our approach,  \algname{FeDABoost}, treats clients as weak learners. Instead of weighting weak classifiers, it dynamically weights client updates based on their local performance. Clients with lower error rates receive higher $\alpha$ values, increasing their influence on the global model. Conversely, clients with higher error rates receive lower $\alpha$ values, reducing their contribution and helping to mitigate the risk of noise or bias from unreliable updates. This approach improves the performance of the global model by down-weighting weak clients in the presence of non-IID data.

In each global round \(e\), let there be \(m\) participating clients. Each client \( j \) trains a simple neural network (NN) \( \mu_j^e \) on its local dataset \( D_j \). Drawing from \eqref{Eq:weight-update-factor}, we define the weighting factor \( \alpha \) for client \( j \) in the \(e\)th training round~as:

\begin{equation}
    \label{Eq:weight-update-factor-fedaboost}
    \alpha_j^e= \ln\left(\frac{1 - \mathcal{E}_j^e}{\mathcal{E}_j^e}\right) + \ln(C_j - 1),
\end{equation}

\noindent
where $C_j$ denotes the number of classes handled by client $j$. Unlike \algname{SAMME}, where \(\alpha\) is based on the error of the weak classifier on a common dataset, here \(\mathcal{E}_j^e\) denotes the error of client \(j\)'s local model ($\mu_j^e$) on its own validation set drawn from \(D_j\). The value \(\alpha_j^e\) is positive only when \(1 - \mathcal{E}_j^e > 1 / C_j\), meaning only clients whose performance exceeds random guessing are included in the aggregation. Clients with non-positive \(\alpha_j^e\) values are ignored, as their updates are likely to degrade the global model.

Crucially, as the number of classes increases, so does the acceptable error threshold for inclusion. This design choice is beneficial for FL settings with non-IID data, as it enables clients with moderate performance to contribute positively by leveraging the ensemble effect. However, including too many weak client models can still negatively impact performance, highlighting the importance of selective weighting and client filtering.

When \(\mathcal{E}_j^e\) approaches 0 or 1, \(\alpha_j^e\) tends toward infinity, causing excessive influence from the local model of that client. Such extreme weighting can adversely impact the aggregation process, introducing overfitting or bias into the global model. To prevent this, \algname{FeDABoost} clips \(\mathcal{E}_j^e\) to the range \([\epsilon, 1 - \epsilon]\), where \(\epsilon\) is a small constant (e.g., \(10^{-6}\)). 

This clipping approach ensures that no individual client disproportionately dominates the global model, while preserving the core principle of \algname{FeDABoost}, which is to amplify contributions from strong clients while controlling the impact from weaker ones. 

Each client computes \(\alpha_j^e\) locally and communicates it alongside its model update \(\mu_j^e\) to the server. The server then aggregates the global model \(\mathcal{M}^{e+1}\) as:

\begin{equation}
    \label{Eq:fedaboost_avg}
    \mathcal{M}^{e+1} = \frac{\sum_{j=1}^{m} \alpha_j^e\mu_j^e}{\sum_{j=1}^{m} \alpha_j^e}\DotPunct
\end{equation}

\subsection{\algname{FeDABoost} Boosting Mechanism}
\label{sec:boosting}

We utilize a weight calculation mechanism inspired by the \algname{SAMME} algorithm to boost the training of underperforming clients in the FL setup. In \algname{SAMME}, the weights for each classifier (\(f_j\)) at iteration ``$e$'' is updated using the following equation: \(w_j^{e} = w_j^{e-1} \cdot \exp\left( \alpha_j \cdot I(f_{j} \textit{ Performance} \right)\)), where \(\alpha_j\) is based on the \(f_j\)'s error rate~\eqref{Eq:weight-update-factor}. The term \(I\) is an indicator function that returns to 0 when \(f_j\)'s meets a defined performance threshold.

A common challenge in \algname{AdaBoost} and \algname{SAMME} is the rapid increase in weights due to the exponential factor of \(\alpha\), which can lead to overfitting. While \algname{SAMME} somewhat eases this issue by integrating the number of classes into the \(\alpha\) calculation as shown in~\eqref{Eq:weight-update-factor}, it does not completely resolve it. To address this, we incorporate a learning rate \(\eta \in [0,1]\) into the weight update mechanism~\cite{hastie2009boosting,scikit-learn_adaboost_multiclass}. This adjustment mitigates the steep increase in weights during the later stages of training, ultimately resulting in a more stable and generalizable model performance. 

The weights are calculated by~\eqref{Eq:weight}:

\begin{equation}
    \label{Eq:weight}
     w_j^{e} = w_j^{e-1}\exp\left( -\eta\alpha_jI(\mu_{j} \textit{ Performance}) \right),
\end{equation}

\noindent The negative sign inverts the influence of \(\alpha_j\), ensuring that clients with higher error rates (and thus lower \(\alpha_j\)) receive increased weights: clients with poorer performance receive higher weights, encouraging their improvement during training.

In the initial global training round, all clients are assigned equal weights, represented as \(w_j = {1}/{m_0}\), for \(j=1,2,\ldots, m_0\), where \(m_0\) denotes the number of clients that participate in the initial global round. After each global round, weights are updated using~\eqref{Eq:weight}, with \(\alpha_j\) derived from the error rate of client \(j\)'s local model~\eqref{Eq:weight-update-factor}. 
\(I(\mu_{j}\textit{Performance})\) is the indicator function that equals \(0\) if the \(\mu_j\)'s performance meets a predetermined accuracy threshold, which should be established empirically. This means that once the client model achieves the specified accuracy, the algorithm will no longer boost the client's model training.

Subsequently, \algname{FeDABoost} utilizes these weights to adjust the \(\gamma\) values in the focal loss~\eqref{Eq:focal_loss} function. At each global round, \(\gamma\) is incrementally increased by the updated weight, enabling the model to emphasize harder samples during training. The \(\gamma\) value is constrained to the range [0, 5] as suggested in~\cite{lin2017focal}. As we assume static data across rounds, we do not update the class imbalance parameter \(\beta\); it is set empirically and remains fixed. Finally, weight updates are computed only for clients that actively participate in each global training round.

\subsection{The proposed \algname{FeDABoost} algorithm}
\label{sec:fedaboost}

The \algname{FeDABoost} algorithm consists of two primary phases: {\it Initialization} and {\it Iteration}.
In the \textit{Initialization} phase, \algname{FeDABoost} initializes the global model, denoted as $\mathcal{M}^e$ at iteration $e=0$ with random parameters. Since each client model serves as a weak learner in \algname{FeDABoost}, a simple NN architecture is particularly well-suited for a global model. The client weights are also set to be equal at the initial round, with each client \(i\) is assigned an initial weight \(w^0_i = {1}/{m_0}\), where \(i=1,2,\ldots, k\) and \(m_0\) is the number of clients participate in the initial training round and \(k\) is the total clients in the federated setup.

At {\it iteration} of global round $e$, the weights $\bigl\{ w^e_i \bigr\}_{i=1}^{k}$ and the global model $\mathcal{M}^e$ are shared by a subset of clients \(S_e \subseteq \{1, 2, \ldots, k\}\), where \(|S_e| = m_e\), who participate in the next training round \((e+1)\). Upon receiving \(\mathcal{M}^{e}\), client \(j\), for $j=1,2,\ldots, m_e$, computes the error rate (\(\mathcal{E}_j^{e}\)) of $\mathcal{M}^e$  for its local data (\(\mathcal{D}_{j}\)). Client \(j\) then calculates \(\alpha_j^{e}\) using~\eqref{Eq:weight-update-factor-fedaboost} and updates the weight (\(w_j^e\)) using~\eqref{Eq:weight}. Later, the client \(j\), trains $\mathcal{M}^e$ several local iterations with \(D_j\) leading to \(\mu_j^{e}\). During the training process, the client’s weight \(w_j^{e}\) boosts the training, as explained in Section~\ref{sec:boosting}. After completing the local training, the new error rate \(\mathcal{E}_j^{e}\) and new \(\alpha_j^{e}\) are computed for the model \(\mu_j^{e}\), and both \(\alpha_j^{e}\) and the trained local model \(\mu_j^{e}\) are sent to the central server. Upon receiving updates from all the clients, a new global model is formed using~\eqref{Eq:fedaboost_avg}. The global model $\mathcal{M}^{e+1}$ is then sent back to all clients. The iteration process continues until the global model converges, as described in Algorithm~\ref{alg:FeDaBoost}.

\begin{algorithm}[ht!]
    \caption{\algname{FeDABoost}.}
    \label{alg:FeDaBoost}
    \begin{algorithmic}[1]
        \Procedure{Server-Side}{}
        \State Initialize $\mathcal{M}^1$ and weights $w_i = 1/k$ for $i = 1, \ldots, k$
        \For {$e=1,2, \ldots, E$}
            \State Set \(S_e \subseteq \{1, 2, \ldots, k\}, \quad |S_e| = m\)
            \For{each client $j \in S_e$ \textbf{in parallel}}
                \State $\mu_j^{e}$, $\alpha_j^{e}$  $\gets$ \Call{Client-Update}{j, $\mathcal{M}^{e-1}$}
            \EndFor
            \State Aggregate $\{ \mu_j^{e}, \alpha_j^{e} \}$ to compute $\mathcal{M}^{e+1}$ via~\eqref{Eq:fedaboost_avg}
        \EndFor
        \EndProcedure
        \Procedure{Client-Update} {Client j, Global Model $\mathcal{M}$}
        \State Compute $\mathcal{E}_j$ (error rate of $\mathcal{M}$), then calculate $\alpha_j$ using~\eqref{Eq:weight-update-factor}, and $w_j$ using~\eqref{Eq:weight}
        \State $\mu_j \gets \text{Train } \mathcal{M}$ on $\mathcal{D}_j$ per local round, with loss influenced by $w_j$
        \State Recompute $\mathcal{E}_j$ and update $\alpha_j$
        \State \Return $\mu_j,  \alpha_j$ 
        \EndProcedure
    \end{algorithmic} 
\end{algorithm}

\subsection{\algname{FeDABoost} Fairness and Convergence}

The \algname{FeDABoost} achieves {\bf convergence towards fairness and performance} (Section~\ref{sec:problem}) by integrating two complementary mechanisms: performance-aware aggregation (Section~\ref{sec:aggregation}) and an adaptive, client-specific boosting mechanism (Section~\ref{sec:boosting}). This ensures that at each global round $e$, the global model loss \(\bar{\ell}(\mathcal{M}^e)\) decreases while the variance in client performance \(\mathrm{Var}(\varphi(\mathcal{M}^e))\) also reduces, eventually stabilizing as $e$ increases.

The aggregation weights $\alpha^e_j$, based on client error rates, assign more influence to better performing clients during model aggregation:
\(\bar{\ell}(\mathcal{M}^{e+1}) \leq \bar{\ell}\left( \frac{\sum_j \alpha^e_j \mu^e_j}{\sum_j \alpha^e_j} \right) < \bar{\ell}(\mathcal{M}^e)\). Concurrently, underperformed clients are dynamically boosted during local training. Specifically, the focal loss parameter $\gamma$ is adaptively increased according to each client's prior performance~\eqref{Eq:focal_loss}, enabling these clients to focus more on harder examples and accelerate loss reduction. As a result, for most clients the following is satisfied:
\(
\ell_j(\mu_j^{e}) \ll \ell_j(\mathcal{M}^{e-1}).
\) 

This dual mechanism reduces performance disparities: as struggling clients improve faster, their performance approaches the average $\varphi(\mathcal{M}^e)$, resulting in decreasing variance across rounds:
\(
\mathrm{Var}(\varphi(\mathcal{M}^{e+1})) \ll \mathrm{Var}(\varphi(\mathcal{M}^{e})).
\)
\noindent
Finally, the parameter $\eta$ handles the sensitivity of the boosting adjustment. When \(\eta\) is close to 1, it converges faster but may lead to instability. Conversely, when \(\eta\) is close to 0, convergence slows down and the boosting effect is reduced.

\section{Experimental Setup}

The \algname{FeDABoost} algorithm is evaluated on three datasets: \textbf{MNIST}, \textbf{\dataset{FEMNIST}}, and \textbf{CIFAR10}\footnote{Dataset links: {MNIST}: \url{http://yann.lecun.com/exdb/mnist}, {\dataset{FEMNIST}}: \url{https://leaf.cmu.edu}, {CIFAR-10}: \url{https://www.cs.toronto.edu/~kriz/cifar.html}}. For \dataset{MNIST} and  \dataset{CIFAR10}, we simulate non-IID conditions using Dirichlet data partitioning~\cite{lin2020ensemble}, with concentration parameters set to 0.2 and 0.4 respectively. \dataset{FEMNIST} is inherently non-IID, containing user-annotated handwriting data distributed across individual writers.

We compare \algname{FeDABoost} with two FL baselines: \algname{FedAvg}~\cite{mcmahan2017communication} and \algname{Ditto}~\cite{li2021Ditto}. \algname{FedAvg} aggregates local model updates by weighting them based on dataset size. In contrast, \algname{Ditto} maintains both global and personalized local models, incorporating a regularization term (\(\mathcal{\lambda}\)) that controls closeness between them.

Three experiments were conducted. In the first experiment (\textbf{Ex.1}), we evaluated \algname{FeDABoost} on the MNIST dataset in comparison with \algname{FedAvg}. Each communication round involved randomly selected 30\% of clients. To analyze the contribution of different components, we performed an ablation study using a variant of \algname{FeDABoost} that utilizes only the alpha-based aggregation and excludes the boosting mechanism. All models were optimized using stochastic gradient descent (SGD) with shared hyperparameters, which were first tuned using \algname{FedAvg} and reused for all other algorithms to ensure a controlled comparison. The local model architecture is a fully connected NN with one hidden layer. 

In the second experiment (\textbf{Ex.2}), we used the \dataset{FEMNIST} dataset to compare \algname{FeDABoost} with both \algname{FedAvg} and \algname{Ditto}. We randomly selected 20\% of clients per round and repeated the experiment three times with different client selections for statistical robustness. The model architecture was a lightweight convolutional NN, consisting of a single convolutional layer with batch normalization and max pooling, followed by dropout and a fully connected output layer. For \algname{FedAvg}, we empirically tuned the hyperparameters using SGD. These settings were reused for \algname{FeDABoost} to ensure a fair and controlled comparison. We also evaluated an alternative configuration of \algname{FeDABoost} using the AdamW optimizer. \algname{Ditto} was trained using the same global model architecture and SGD optimizer as \algname{FedAvg}, while the personalized models maintained by each client were updated locally using the Adam optimizer. 

In the third experiment (\textbf{Ex.3}), we evaluated \algname{FeDABoost} on the \dataset{CIFAR10} dataset in comparison with \algname{FedAvg} and \algname{Ditto}. This experiment was designed to investigate the impact of varying client participation rates on model performance. We conducted training runs using three different client participation fractions: 20\%, 40\%, and 60\%, with clients randomly selected in each communication round. To ensure a fair and controlled comparison, we reused the SGD-tuned hyperparameters originally optimized for \algname{FedAvg} across all methods. \algname{Ditto} was configured consistently with {\bf Ex.2}. The global model was trained using the same settings as \algname{FedAvg}, while the personalized models maintained by each client were updated locally using the Adam optimizer. The local model architecture was a convolutional NN consisting of two convolutional layers with group normalization, followed by max pooling, dropout, and two fully connected layers. This setup allowed us to assess how different levels of client availability influence the relative effectiveness of \algname{FeDABoost}.

In all three experiments, we assessed the global model at each training round, using the loss, $F1$ score, and accuracy calculated from the global validation set, which comprises 20\% of unseen data from each client. In all methods, the parameter \(\mathcal{\beta}\) in~\eqref{Eq:focal_loss} was set to 1.

\section{Evaluation and Results}
\begin{figure}[!htp]
    \centering        \includegraphics[width=0.30\columnwidth]{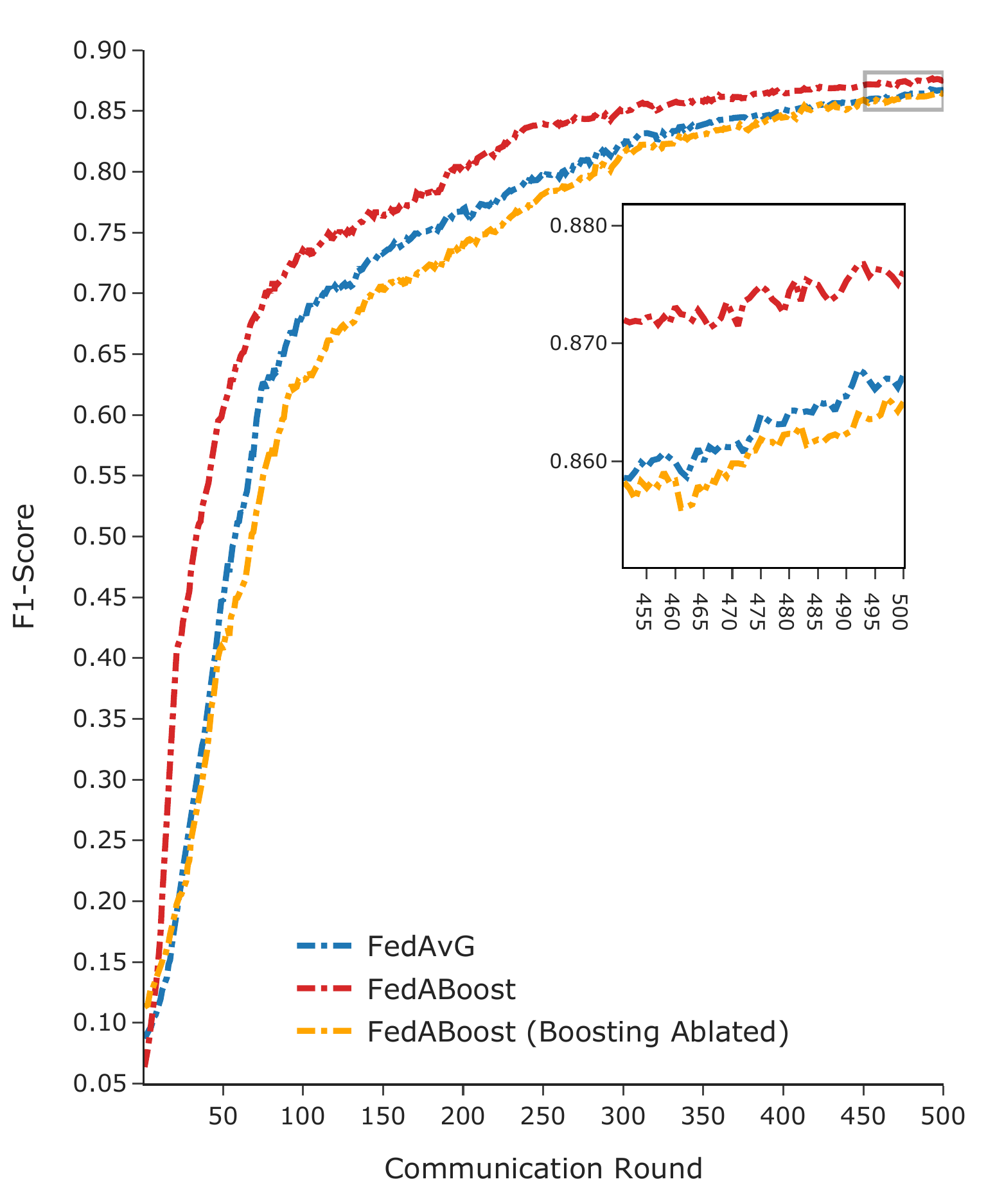} 
        \includegraphics[width=0.30\columnwidth]{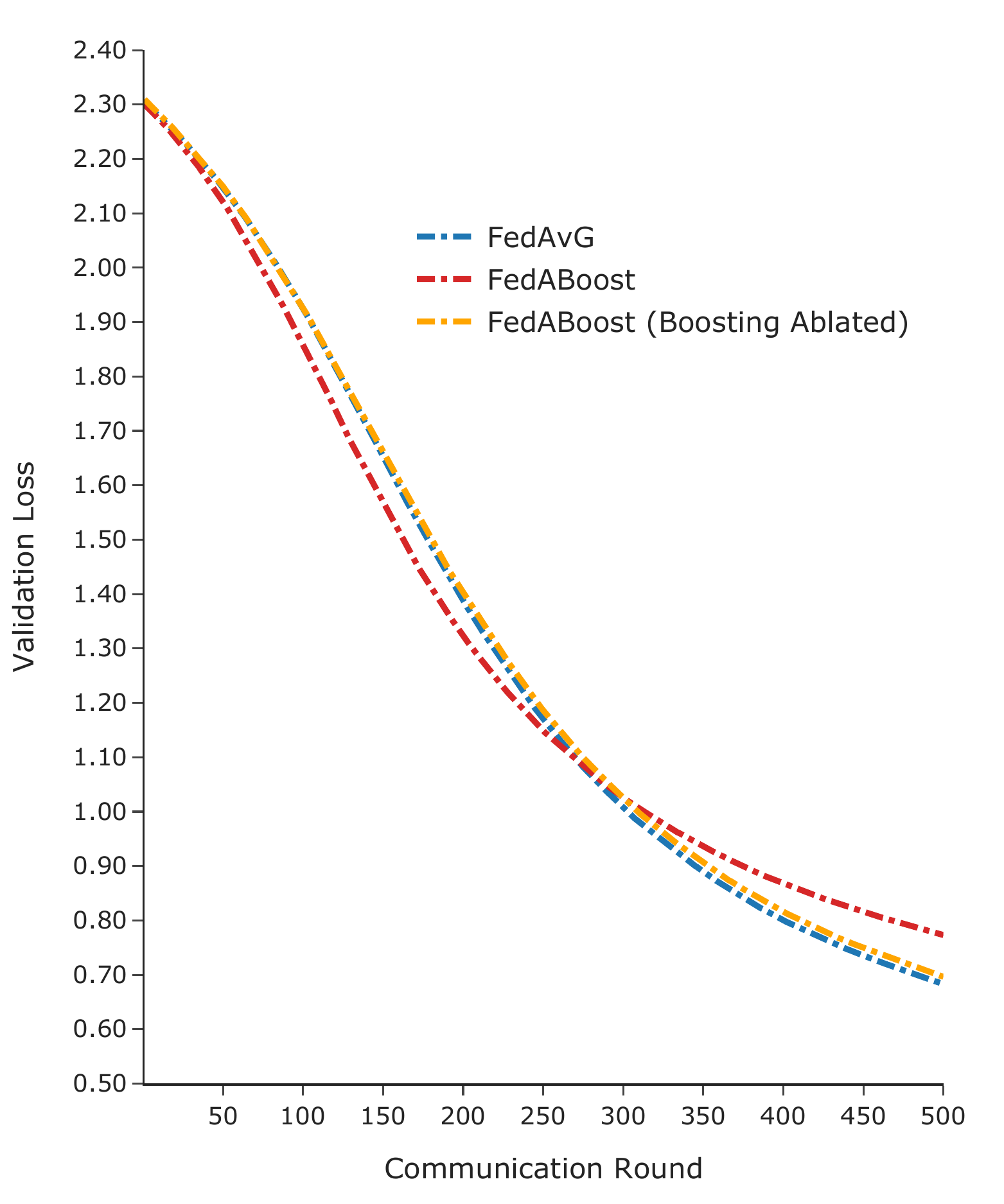} 
    \caption{\textbf{Ex.1: MNIST.} All models are trained with SGD (learning rate = \(1 \times 10^{-3}\), batch size = 32, weight decay = \(1 \times 10^{-3}\), local epochs = 5); \algname{FeDABoost} $\mathcal{\eta}=0.01$ and error threshold = 0.3. Total clients: 264.}
    \label{fig:mnist-results}
\end{figure}

The results of {\bf Ex.1} are presented in Figure~\ref{fig:mnist-results}. We observe that \algname{FeDABoost} consistently outperforms \algname{FedAvg}. Specifically, \algname{FeDABoost} achieves an $F1$ score of approximately 0.88 in convergence, while \algname{FedAvg} saturates around 0.87. The comparison between \algname{FedABoost} and its ablated variant (without boosting) clearly demonstrates the critical role of boosting in enhancing model performance. This indicates that boosting contributes to addressing non-IID data challenges and client heterogeneity in FL. Another critical observation is the communication efficiency demonstrated by \algname{FeDABoost}, as it consistently reaches higher $F1$ scores in fewer communication rounds compared to \algname{FedAvg}, which indicates a reduction in communication overhead for a given performance target. This is particularly valuable in FL environments, where communication cost is a bottleneck. It is also noted that the validation loss curves exhibit a temporary crossing around communication round 250, likely due to the dynamic adjustment of the focal loss \(\gamma\) parameter in \algname{FeDABoost}. 

As  \(\gamma\) increases, the model prioritizes harder samples, which eventually increases loss without significantly impacting the $F1$ score, as prediction accuracy remains stable.

\begin{figure}[!htp]
    \centering
    {
        \includegraphics[width=0.30\textwidth]{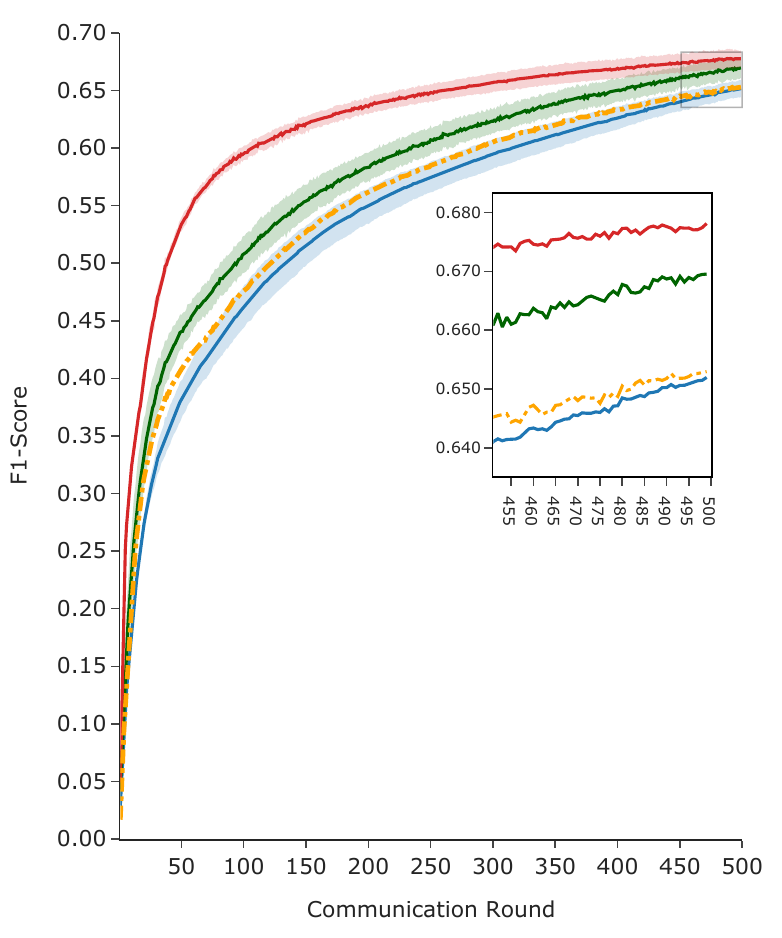}
        \includegraphics[width=0.30\textwidth]{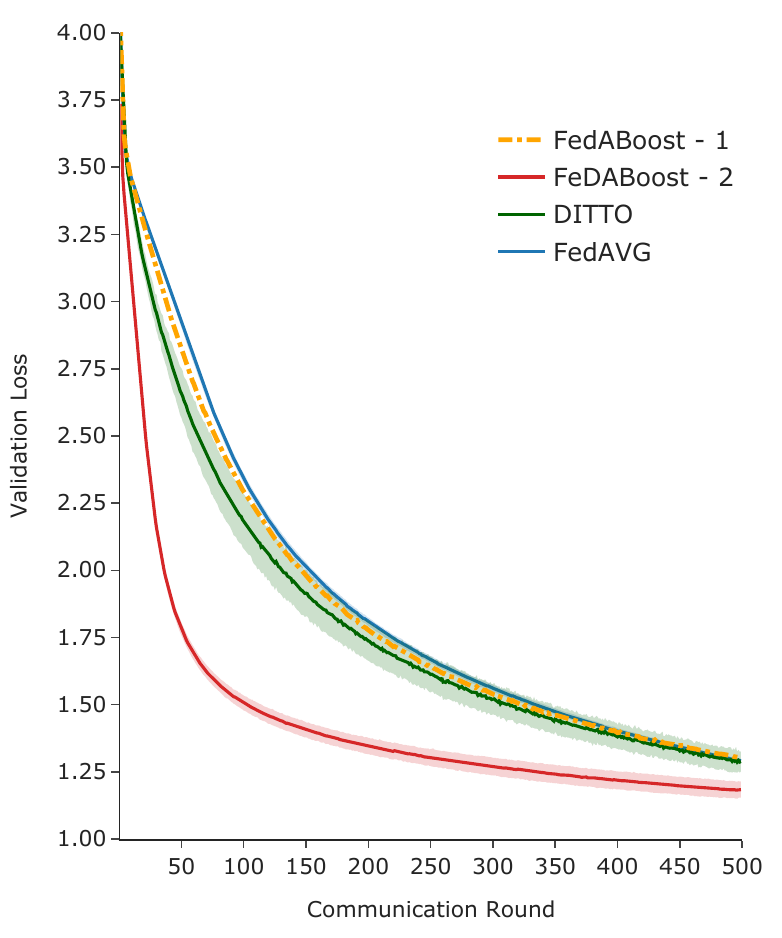}
        \label{fig:femnist-f1score}
    }
    \caption{\textbf{Ex.2: \dataset{FEMNIST}.} All models are trained for 5 local epochs. The global models of \algname{FedAvg}, \algname{FedABoost}-1, and \algname{Ditto} use SGD (learning rate = $10^{-3}$, batch size = 64, weight decay = $5 \times 10^{-4}$). \algname{Ditto}'s personalized models use Adam (learning rate = $10^{-3}$, batch size = 64, weight decay = $5 \times 10^{-4}$, $\lambda = 0.1$). \algname{FedABoost}-2 uses AdamW (learning rate = $2 \times 10^{-4}$, batch size = 64, weight decay = $10^{-6}$). \algname{FedABoost} uses $\eta = 0.01$ and error threshold = 0.5. Total clients: 3,550.}
    \label{fig:femnist-results}
\end{figure}

The results of {\bf Ex.2} are shown in Figure~\ref{fig:femnist-results}. \algname{FedABoost}-1 consistently achieves higher $F1$ scores across communication rounds when compared to \algname{FedAvg}. This highlights the effectiveness of its dynamic boosting and aggregation mechanisms. However, \algname{FedABoost}-1’s dependence limits optimization efficiency due to sensitivity to learning rate and lack of adaptivity of SGD. In contrast, \algname{FedABoost}-2, using AdamW, significantly outperforms all baselines, achieving faster and more stable convergence. This gain is likely from AdamW’s adaptive learning rate and decoupled weight decay, which better accommodate the variability introduced by \algname{FedABoost}’s $\alpha$-based aggregation and dynamic $\gamma$ adjustment in focal loss. We have also experimented with both optimizers across algorithms and found that SGD was more stable with \algname{FedAvg}, while AdamW worked better with \algname{FedABoost}, resulting in a more stable learning curve.

\algname{Ditto} trained its global model using SGD, like \algname{FedAvg}, while maintaining personalized models for each client using Adam. We can see that \algname{Ditto} consistently outperforms \algname{FedABoost}-1. However, it does not surpass \algname{FedABoost}-2, which leverages AdamW in a fully collaborative setting with dynamic client weighting and loss modulation. This suggests that while \algname{Ditto} benefits from personalization through local adaptivity, it lacks the collective optimization enhancements introduced by \algname{FedABoost}'s $\alpha$-based aggregation and dynamic $\gamma$ adjustment. Moreover, \algname{Ditto} requires additional memory due to dual model maintenance, which can pose challenges for some clients, especially those with constrained resources.

\begin{figure}[!htp]
    \centering        \includegraphics[width=0.48\columnwidth]{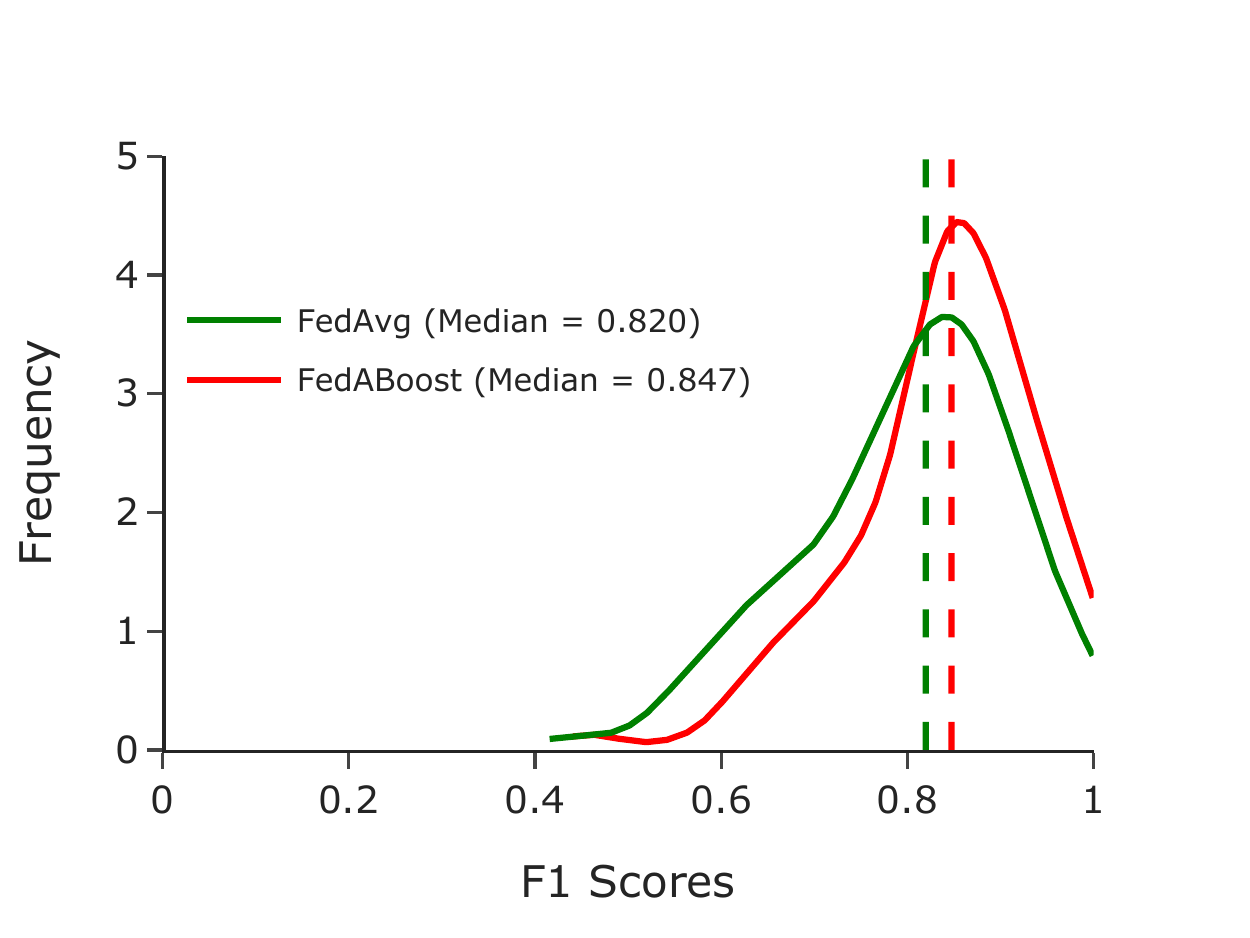} 
        \includegraphics[width=0.48\columnwidth]{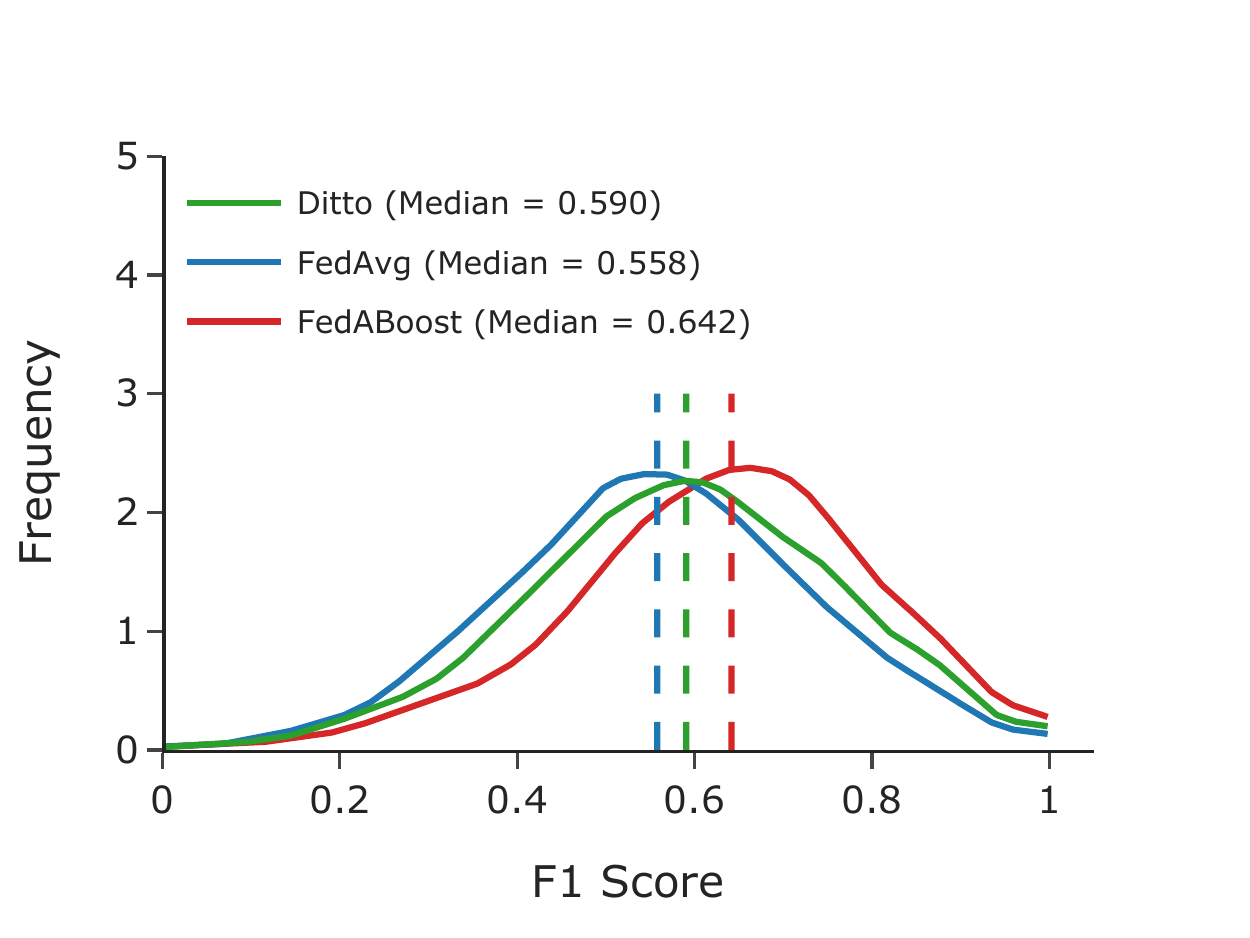} 
    \caption{
    $F1$ score distributions: Left – \dataset{MNIST}; Right – \dataset{FEMNIST}.
    }
    \label{fig:fedaboost-results-performance-farness}
\end{figure}

In {\bf Ex.1} and {\bf Ex.2}, the distribution of $F1$ scores across all clients is presented in Figure~\ref{fig:fedaboost-results-performance-farness}. In both \dataset{MNIST} (left) and \dataset{FEMNIST} (right), \algname{FedABoost} produces a right-shifted, more concentrated distribution compared to baselines, indicating improved performance and consistency. It achieves the highest median $F1$ scores in both cases; 0.852 on MNIST (vs. 0.813 for \algname{FedAvg}) and 0.652 on \dataset{FEMNIST} (vs. 0.566 for \algname{Ditto} and 0.558 for \algname{FedAvg}).

To quantify fairness improvements, we calculated the variance of $F1$ scores across clients over the convergence windows (global rounds 245–255 for \dataset{MNIST}, and global rounds 205–210 for \dataset{FEMNIST}). On \textsc{MNIST}, \algname{FedABoost} achieves lower variance of 0.0103 with a 95\% confidence interval (CI) of [0.0102, 0.0104], reducing variance by 24.4\% compared to \algname{FedAvg} (0.0137; 95\% CI of [0.0135, 0.0138]). On \textsc{FEMNIST}, \algname{FedABoost} achieves an average variance of 0.0279 with a 95\% CI of [0.0278, 0.0280], representing a 5.88\% reduction compared to \algname{FedAvg} (0.0296; CI=[0.0295, 0.0297]) and an 11.87\% reduction over \algname{Ditto} (0.0317; CI=[0.0313, 0.0320]). These results, supported by non-overlapping confidence intervals, confirm that \algname{FedABoost} not only enhances the performance but also improves fairness across clients, fulfilling objective defined in~\eqref{eq:objective}.

The results of {\bf Ex.3}, shown in Table~\ref{tab:algorithm_comparison}, indicate that \algname{FeDABoost} outperforms both \algname{FedAvg} and \algname{Ditto} across all client fractions in terms of $F1$ score, with the most notable margin at 20\% participation (0.716 vs. 0.694 and 0.689). As can be observed, the performance gap is not as significant as in {\bf Ex. 1} and {\bf Ex. 2}, where the data were highly non-IID. Note that the \dataset{CIFAR10} dataset client fraction was performed with a concentration parameter of 0.4, resulting in moderately non-IID data. \algname{FedABoost} also shows lower validation loss at lower participation levels, indicating better generalization under constrained settings. This aligns with the design of \algname{FedABoost}, which enhances underrepresented clients. This effect is most beneficial when fewer clients participate per round, allowing their influence to be more effectively integrated into the global model.

\begin{table}[ht]
\small
\centering
\caption{\textbf{Ex.3: \dataset{CIFAR10}.} All global models of \algname{FedAvg}, \algname{FedABoost}, and \algname{Ditto} are trained for 10 local epochs using SGD (learning rate = $10^{-2}$; batch size = 32). \algname{Ditto}'s personalized models are trained with SGD (learning rate = $10^{-3}$, batch size = 32, $\lambda = 0.1$). \algname{FedABoost} uses $\eta = 0.002$, error threshold 0.4. Total clients: 196. The global model converged in approximately 60–70 rounds with 20\% of the data, 35–45 rounds with 40\%, and 25–30 rounds with 60\%.}

\renewcommand{\arraystretch}{0.9} 
\setlength{\tabcolsep}{3pt} 
\begin{tabular}{|l|ccc|ccc|}
\hline
\multirow{2}{*}{} & \multicolumn{3}{c|}{\textbf{F1 Score}} &  \multicolumn{3}{c|}{\textbf{Validation Loss}} \\ \cline{2-7}
& \textbf{20\%}   & \textbf{40\%}   & \textbf{60\%}   & \textbf{20\%}   & \textbf{40\%}   & \textbf{60\%}   \\ \hline
\algname{FeDABoost} &  $0.716$  &  $0.685$  & $0.677$ &   $0.992$ & $1.051$ & $1.078$ \\ \hline
\algname{FedAvg} & $0.694$ & $0.679$ & $0.675$ & $1.198$ & $1.149$ & $1.156$ \\ \hline
\algname{Ditto} & $0.689$ & $0.672$ & $0.642$  & $1.199$ & $1.208$ & $1.355$ \\ \hline
\end{tabular}
\label{tab:algorithm_comparison}
\end{table}

Our experiments have revealed that \algname{FeDABoost} is more efficient in non-IID settings than in IID, compared to the two baselines. Furthermore, \algname{FeDABoost} has demonstrated more stable behavior during federated training than \algname{FedAvg}, achieving a faster loss reduction.

\section{Literature Review}

The model aggregation techniques in FL can be categorized into two main types: parameter-based aggregation and output-based aggregation~\cite{QI2024}. This classification is determined by the nature of the objects being aggregated. Parameter-based aggregation~\cite{wang2024fedave} focuses on the combination of trainable parameters from local models, including weight parameters and gradients from deep NNs. In contrast, output-based aggregation underlines the aggregation of model representations, such as output logits or compressed sketches. In~\cite{nabavirazavi2024enhancing}, the authors propose three new aggregation \textit{functions-Switch}, \textit{Layered-Switch}, and \textit{Weighted FedAvg} to enhance robustness against model poisoning attacks.

\textbf{Collaboration Fairness}: CFFL~\cite{lyu2020collaborative} and RFFL~\cite{xu2020reputation} promote collaborative fairness through a reward mechanism that evaluates client contributions and iteratively adjusts rewards based on gradient updates. Qiuxian et al.~\cite{qiuxian2024secure} propose a fairness mechanism using rewards for improvements in clients' model performance and penalties for deviations from the global model. FedAVE~\cite{wang2024fedave} uses an adaptive reputation calculation module to evaluate clients' reputations based on their local model performance and data similarity to a validation set. A dynamic gradient reward distribution module then allocates rewards based on these reputations, ensuring that more valuable contributions receive larger rewards. Wang et al.~\cite{wang2024fedsac} discuss the disadvantages of approaches that achieve fairness by adjusting clients' gradients~\cite{lyu2020collaborative,wang2024fedave,xu2020reputation} noting that these methods often fail to maintain consistency across local models and do not adequately address the needs of high-contributing clients. To tackle this issue, the authors propose
\algname{FedSAC}, which dynamically allocates sub-models to each client based on their contributions, rewarding those who contribute more to the learning process with higher-performing sub-models.

\textbf{Performance Fairness}: Zhang et al.~\cite{zhang2020fairfl}
propose the FairFL framework, which uses deep multi-agent reinforcement learning alongside a secure information aggregation protocol to optimize both accuracy and fairness while ensuring privacy. FairFed~\cite{ezzeldin2023fairfed} algorithm improves fairness in FL by adjusting the model aggregation weights based on local fairness, which assesses the model performance across different demographic groups within a client's dataset. Ditto~\cite{li2021Ditto} achieves fairness by creating personalized models by combining a global model and local objectives for each client, which helps address the variability in data across clients. 

\section{Conclusion and Future Directions} 
In this work, we have introduced \algname{FeDABoost}, a novel FL framework that enhances global model quality by fairly boosting clients based on local performance. Evaluations on MNIST, FEMNIST, and CIFAR10 data demonstrate that \algname{FeDABoost} generally outperforms \algname{FedAvg} and \algname{Ditto}, particularly in non-IID settings and those with limited client participation. Furthermore, results suggest that \algname{FeDABoost} is particularly well suited to cross-silo FL settings, where fairness and interpretability of client contributions are critical.

Despite its promise, \algname{FeDABoost} shows sensitivity to its hyperparameters and can be fragile in certain settings. 
In our future work, therefore, we plan to explore alternative boosting strategies beyond focal loss, and incorporate adaptive mechanisms such as error-threshold and $\eta$ scheduling, and robust optimizer tuning to further improve stability and generalization across diverse scenarios.

\bibliographystyle{splncs04}

\end{document}